\documentclass[a4paper,10pt]{article}
\usepackage[utf8]{inputenc}
\usepackage{amsfonts}
\usepackage{amsmath}

\usepackage{algorithm}
\usepackage{algpseudocode}
\usepackage{graphicx}
\usepackage{indentfirst}
\usepackage{subfig}
\newcommand{\argmax}{\operatornamewithlimits{arg\,max}}
\newcommand{\argmin}{\operatornamewithlimits{arg\,min}}

\begin{document}

\title{Linear Function Approximation as a Computationally Efficient Method to solve Classical Reinforcement Learning Challenges}
\author{Hari Srikanth \\\\
}
\date{July 30, 2022}
\maketitle

\begin{abstract}

Neural Network based approximations of the Value function make up the core of leading Policy Based methods such as Trust Regional Policy Optimization (TRPO) and Proximal Policy Optimization (PPO).  While this adds significant value when dealing with very complex environments, we note that in sufficiently low State and action space environments, a computationally expensive Neural Network architecture offers marginal improvement over simpler Value approximation methods. We present an implementation of Natural Actor Critic algorithms with actor updates through Natural Policy Gradient methods. This paper proposes that Natural Policy Gradient (NPG) methods with Linear Function Approximation as a paradigm for value approximation may surpass the performance and speed of Neural Network based models such as TRPO and PPO within these environments. Over Reinforcement Learning benchmarks Cart Pole and Acrobot, we observe that our algorithm trains much faster than complex neural network architectures, and obtains an equivalent or greater result. This allows us to recommend the use of NPG methods with Linear Function Approximation over TRPO and PPO for both traditional and sparse reward low dimensional problems. 

\end{abstract}

\newpage

\section{Introduction}

    Reinforcement Learning (RL) is a paradigm where an agent seeks to maximize the reward it gains through refining its policy. At each timestep t, our agent observes the environmental State, and according to policy $\pi$ it takes some action. This action changes the environmental State and returns some reward, and this is used to retrain the policy (Figure 1). This simple procedure has applications in numerous fields, such as developing Self Driving Cars \cite{kiran2021deep} or training a robot to solve a rubix cube \cite{openai2019solving}.

    \begin{figure}[hbt!]
    \centering
    \includegraphics[width=7cm]{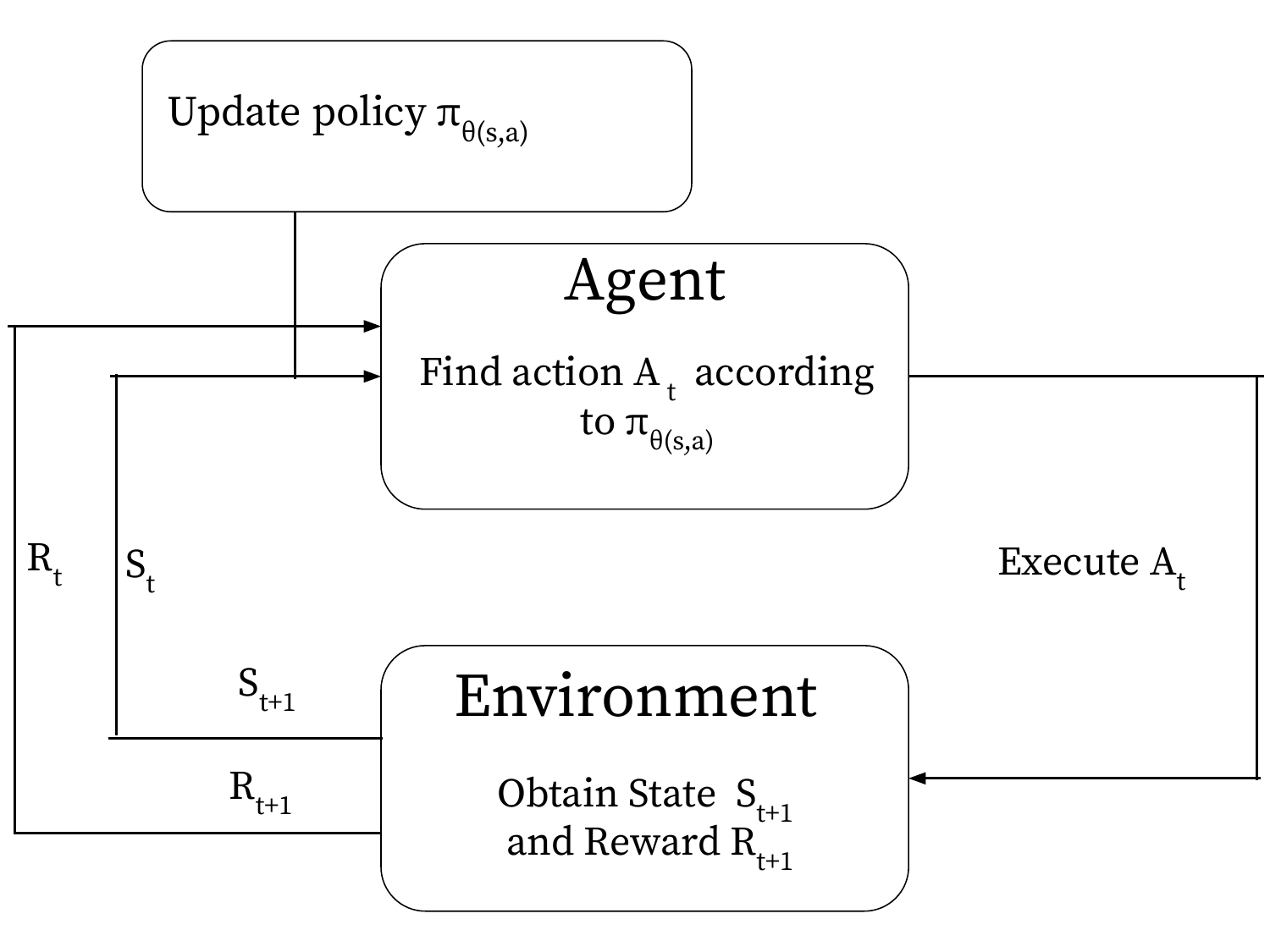}
    \caption{Simplified RL procedure}%
    \label{fig:example}%
    \end{figure}

    The objective of our algorithm is to maximize the cumulative reward the agent obtains over a number of timesteps $t$. As such we hope to determine the policy that results in the maximum expected cumulative reward. There exist 2 main branches of RL to determine find this policy: Value Based Methods (such as Q-Learing) and Policy Based Methods (such as Actor Critic). We will focus on Policy-based methods due to their more stable performance and predictable convergence. One of the most well researched Policy based methods is Actor Critic, which we can describe as follows: Actor critic methods feature an actor (which updates policy through gradients) and a critic (which estimates the value function) \cite{BHATNAGAR20092471, PETERS20081180, khodadadian2021finite}. These elements work together in tandem to develop a system that guarantees convergence for linear and non-linear(neural network) approximations of the value function, along with reducing variance compared to standard Policy based Methods. Actor critic methods are used in many fields, such as Robotics \cite{haarnoja2019soft} and Network control \cite{xu2021graphembedded}. There also is a variation of actor critic, called Natural Actor Critic. Instead of using standard gradient descent to update the policy during the actor step, Natural Actor Critic utilizes natural gradient descent. This results in a much quicker convergence to the optimal policy.
    
    Currently, most implementations of Policy Based methods utilize complex Neural Networks to estimate the value function. This is what lies at the core of industry standard algorithms such as Trust Regional Policy Optimization or Proximal Policy Optimization. Although Neural Network based methods such as Trust Region Policy Optimization (TRPO) \cite{schulman2017trust} and Proximal Policy Optimization (PPO) \cite{schulman2017proximal}are widely used for the vast majority of RL Applications, they still retain the drawbacks of Neural Network type architectures. The most notable of these are the computationally expensive nature of Neural Networks and the difficulty for development. In addition, we theorize that there are a multitude of scenarios where the nature of the problem does not necessitate Neural Networks. With this in mind, we aim to use Linear Function Approximation (LFA) as a simpler paradigm to implement RL algorithms. We aim to accomplish this through the use of a Natural Policy Gradient (NPG) algorithm that utilizes LFA (LFA-NPG). We evaluate these methods with regards to three aspects:
    \begin{itemize}
        \item Performance (Reward vs. number of policy iterations): In our testing, we observe that LFA-NPG will match the performance of PPO and TRPO in standard applications, and outperform these algorithms in sparse reward environments.
        \item Speed (Time vs. Reward): We observe that LFA-NPG is noticeably faster than TRPO and PPO in standard applications, and in sparse reward environments it significantly outperforms TRPO, while matching the performance of PPO.
        \item Robustness (Performance over levels of noise): We note that in some environments, LFA-NPG exhibits a higher level of resistance against adversarial noise  than TRPO and PPO.
    \end{itemize}
    
    For the remainder of this paper, we will conduct an analysis of leading Policy based methods, TRPO and PPO. We then outline the structure and design of our LFA-NPG algorithm, and proceed to compare it with TRPO and PPO. We then summarize our findings and draw conclusions.

\section{Literature Review}

    We begin our analysis with an assessment of modern RL. Policy Gradient Methods are a type of RL methods that optimize a parametrized policy through gradient descent. We have decided to focus on Policy Gradient Methods for a few reasons:
\begin{itemize}
    \item Policy gradient methods may utilize the same optimization techniques for States with adversarial noise ($\zeta$)
    \item The same methods may be used to describe discrete and continuous action spaces
    \item Policy Gradient Methods can utilize knowledge of the problem to minimize the parameters required for learning
    \item Policy Gradient Methods may function both with and without a model
\end{itemize}
    Trust Region Policy Optimization (TRPO) and Proximal Policy Optimization (PPO) are the two most researched and implemented Policy Gradient Methods. We will now establish an understanding of the principles behind each algorithm, to develop an intuitive understanding of the strengths and drawbacks of each.
    
    Trust Regional Policy Optimization (TRPO) \cite{schulman2017trust}: TRPO updates each policy by taking the largest possible step possible within some constraint. This constraint is known as the Kullback-Leibler-Divergence \cite{roberts21}, which is analogous to a measure of the distance between probability distributions. To accomplish this large step, TRPO employs complex second order methods to ensure an optimal performance. This is the notable distinction between TRPO and normal policy gradient algorithms, which keep policies relatively close to each other in parameter space. Often, we observe that small differences in parameter space may have significant effect on performance, which requires us to avoid large step sizes with normal policy gradient methods. However, TRPO avoids these pitfalls, which enables it to quickly improve its performance. 

    Proximal Policy Optimization (PPO) \cite{schulman2017proximal}: PPO is a policy gradient method that may define its objective as taking the largest possible policy step within some limiting constraint, similar to TRPO. PPO aims to utilize a set of first-order methods to attain the same results as TRPO within a simpler framework. It differs from TRPO in that it has no constraint, but rather relies on clipping the reward function to remove incentives for policy steps to be sufficiently large.

\section{Purpose}

    Our objective is comprised of three parts:
\begin{enumerate}
    \item Can Linear Function Approximation based Natural Policy Gradient (methods match the performance of State of the art neural network based algorithms such as TRPO and PPO?
    \item Can LFA reach optimal rewards in less time than leading algorithms such as TRPO and PPO?
    \item How do NPG algorithms compare with TRPO and PPO with regards to noise resistance?
\end{enumerate}
    We hope to demonstrate that Natural Policy Gradient Algorithms can match performance and reach rewards in less time than the leading RL algorithms, TRPO and PPO. If we can successfully demonstrate this, then we provide validation for the use of LFA architectures over Neural Networks in many use cases. 

\section{Methodology}

     In order to develop an algorithm that can develop an intelligent agent, we require some framework in order to model the problem. To accomplish this, we turn to the Markov Decision Process (MDP), a mathematically idealized form of the challenge, to which we can make precise theoretical Statements \cite{bartosutton}. We may define the finite MDP as a 5-tuple $(\mathcal{S}, \mathcal{A}, \mathcal{R}, \mathcal{P}, \gamma)$. At each timestep $k$, the agent takes some action $A_k \in \mathcal{A}(s)$ depending on the environment State $S_k \in \mathcal{S}$ and receives some reward $ \mathcal{R}(s_k, a_k) \subset \mathbb R$, and observes some new environmental State $S_{k+1}$ according to transition probability $\mathcal{P}$. This results in a sequence 
     \[
     S_0\underset{\downarrow \mathcal{R}_0}{\rightarrow} A_0\overset{\mathcal{P}(\cdot|S_0,A_0)}{\longrightarrow} S_1\underset{\downarrow R_2}{\rightarrow} A_1 \overset{\mathcal{P}(\cdot|S_1,A_1)}{\longrightarrow} S_2\underset{\downarrow R_2}{\rightarrow}
     A_2\overset{\mathcal{P}(\cdot|S_2,A_2)}{\longrightarrow} S_3\underset{\downarrow R_3}{\rightarrow}...
     \]
    As per the definition of finite MDP, we may take $\mathcal{S}$, $\mathcal{A}$, and $R$ as discrete and finite sets. This allows us to conclude that for each $s^{'} \in \mathcal{S}$ and $r \in \mathcal{R}$, we can determine the probability of those values occurring at timestep t, given the preceding action and State: $$p(s^{'}, r|s,a)  \dot{=}  Pr\{S_k=s^{'}, R_k=r|S_{k-1}=s, A_{k-1}=a\}$$ for all $s^{'}, s \in \mathcal{S}$, $a \in A(s)$, and $r \in \mathcal{R}$. The dynamics of our MDP system can then be described as follows:

    At timestep $k$, observe environmental State $S_k \in \mathcal{S}$. Determine $A_k \in \mathcal{A}$ according to policy $\pi$, $A_k \sim \pi(\cdot |S_k)$.The system then reaches some new State based on the determined transition probabilities $P(S_{k+1} = \cdot|S_k, A_k)$, and returns reward $R(S_k, A_k)$. We seek to maximize our cumulative reward, which we define through our value function. This may be expressed as $$V^{\pi}(s) = \mathbb E[\sum_{k=0}^{\infty} \gamma ^{k}R(S_k, A_k)|S_0 = \mu, A_k \sim \pi (\cdot|S_k)]$$where $\mu$ is an initial distribution over all States. We also define our value function $V^{\pi}(p)=E_{s \sim p}[V^{\pi}(s)]$. Given this, we define the objective of our algorithm as to find the optimal policy $\pi ^*$, such that $$\pi ^{*} \in  \argmax_{\pi \in \Pi} V^{\pi}(\rho)$$ where $\Pi$ is the set of all policies.Having defined our objective, we proceed to determine a system to maximize the reward. Parametrizing the policy with a parameter $\theta$, we aim to find some $\pi_{\theta}$ to maximize $V^{\pi _{\theta}}(\rho)$. We may then redefine our objective as  $\theta ^*$, such that $\theta ^{*} \in  \argmax_{\theta} V^{\pi _{\theta}}(\rho)$. As per our definition of policy, we know $\sum_{a} \pi (a|s) = 1 $ for all $s\in \mathcal{S}$. In order to parametrize our policy $\pi$, we introduce a set of function approximators, $\phi _{s, a} \in \mathbb{R}^2$. We then may define our policy in terms of $\phi$ and $\theta$, as $$\pi_{\theta(a|s)} = \frac{\exp(\phi_{s, a}^\top \theta))}{\sum\limits_a \exp(\phi_{s, a}^\top \theta)}$$
    \begin{algorithm}
\caption{Sampler for: $s,a \sim d_v^\pi$ and unbiased estimate of $Q^\pi(s,a)$}
\label{CHalgorithm}
\begin{algorithmic}[1]
\Require{Starting State-action distribution $\nu$}{}
    \State Sample $s_0, a_0 \sim \nu$
    \State Sample $s,a \sim d_\nu^\pi$, such that at each timestep $h$, with probability $\gamma$, select actions according to $\pi$, else accept $(s_h, a_h)$ as the sample and progress to Step 4.
    \State From $s_h,a_h$, continue to act according to $\pi$ with termination probability of $1-\gamma$. After termination, set ${\hat{Q}}^{\pi(s_h, a_h)}$ as the undiscounted sum of rewards from time h on.
 \end{algorithmic}
   
    \Return{$s_h, a_h$ and $\hat{Q}^\pi(s_h, a_h)$}{}
\end{algorithm}
    
    As such, given $\phi$, we aim to find the optimal vector $\theta$. We plan to optimize $\theta$ via gradient ascent. We update gradient ascent through  $\theta _{k+1} = \theta_{k} + \eta \nabla V^{\pi_\theta}(\rho), \theta \in \mathbb{R}^d$ to find the optimal $\theta^*$. Utilizing the policy gradient theorem, we then find $$\nabla V^{\pi_\theta} = \mathbb{E}\left[Q_{(s,a)}^{\pi_\theta}\nabla_\theta log \pi_\theta(a|s)\right]$$
     $$Q_{s,a}^{\pi} = E\left[\sum_{k= 0}^{\infty} \gamma ^{k}R(S_k, A_k)|S_0 = \mu, A_0 = a, A_k \sim \pi (\cdot|S_k)\right]$$

     Knowing this, we may then estimate $\nabla V^{\pi_{\theta_k}}$, and then update policy through each iteration. To accomplish this, we utilize Natural Actor Critic \cite{agarwal2020theory}. We can break down the actor critic method into two steps:
     \begin{itemize}
         \item Critic: Estimate $\nabla V^{\pi_{\theta_k}}$ with Linear Function Approximation (Algorithm1)
         \item Actor: Update policy with Value-based Natural policy Gradient algorithm (LFA-NPG), utilizing Natural Gradient Descent (Algorithm 2): $$\theta _{k+1} = \theta_{k} + \ H(\theta_k)\nabla V^{\pi_\theta}(p), \theta \in \mathbb{R}^d$$
     \end{itemize}
     
\begin{algorithm}
\caption{Sample-based LFA-NPG for Log-linear Policies}
\label{Calgorithm}
\begin{algorithmic}[1]
\Require{Learning rate $\eta$; Standard Gradient Descent (SGD) Learning Rate $\alpha$; Number of SGD iterations N}{}

    \State Initialize $\theta^0=0$
    \For {t = 0,1,...,T-1} 
        \State Initialize $\omega_0=0$
        \For{n=0,1,...,N-1} 
            \State Call Algorithm 1 to obtain $\hat{Q}(s,a)$ and $s,a \sim d^{(t)}$
            \State Update with SGD: $$\omega_{n+1}=Proj_{\mathcal{W}}(\omega_n -2\alpha(\omega_n \cdot \phi_{s,a}-\hat{Q}(s,a))\phi_{s,a})$$ where $\mathcal{W}=\{w:\|w\|_2 \leq \mathcal{W}\}$
        \EndFor
        \State Set $\hat{w}^{(t)}=\frac{1}{N}\sum_{n=1}^N \omega_n$
        \State Update  $\theta^{(t+1)}=\theta^{(t)}+\eta(\hat{w}^{(t)})$
    \EndFor

\end{algorithmic}
\end{algorithm}

      Natural Gradient Descent: Natural Gradient Descent is a version of gradient descent where each step is multiplied by a Fisher Matrix. In standard Gradient Descent (SGD), we observe that the Gradient of the value function will be small if the predicted distribution is close to the true distribution, and large if the predicted distribution is far from the true distribution. In Natural Gradient Descent, we do not restrict the movement of the parameters in the space, and instead control the movement of the output distribution at each step. We do this by measuring the curve of the Log Likelihood of the probability distribution, or the Fisher Information. This then allows us to focus the direction of convergence directly towards the global minimum, as opposed to SGD, which may not have as direct or swift a convergence. 
      
     We then find our Natural Policy Gradient Algorithm: 
     $$\omega_t \in \argmin_{\omega} E_{s \sim d_p^{\pi_{\theta_k}}, a \sim \pi_{\theta_k}}\left[(Q_{(s,a)}^{\pi_{\theta_k}}-\omega^\top \phi_{s,a})^2 \right]$$
     $$\theta_{k+1} = \theta_k + \eta \omega_k$$
     where  $E[\hat{Q}^\pi(s,a)]=Q^\pi(s,a)$, $\theta \in \mathbb{R}^d$, $\omega \in \mathbb{R}^d, \phi_{s,a} \in \mathbb{R}^d \forall_{s,a}$\\

    An intuitive analysis of LFA-NPG: The agent tests its policy in simulation with termination probability $\gamma$. After either the environment closes or the algorithm terminates, the final State, action, and cumulative reward produced are returned(Algorithm 1). The cumulative reward, State, and action are then used to run SGD on $w$. If the norm of $w$, $\|W\|$ exceeds some limit $\mathcal{W}$, divide $w$ by $\|w\|$. The average $w$ over $N$ iterations is determined, and used to update $\theta$. This occurs for $T$ iterations, after which point we expect to have found the optimal $\theta^*$ that maximizes $V^{\pi _{\theta}}(\rho)$.
    
    We now go on to consider the set of function approximators for our policy, $\phi$. When we parametrize our policy $\pi$, we describe it in terms of $\theta$, the vector we optimize, and $\phi$, the set of function approximators. $\phi$ is defined through $s$ and $a$, where we find $\phi$ as a diagonal $a$ by $a \times s$ matrix, with all nonzero elements as the State $s$. Intuitively, we find that the dimensionality of $\phi$ determines the dimensionality of $\theta$, and as a result of this has control over the performance and speed of LFA-NPG. By modifying the dimensions of the State vector $s$ by adding or removing challenge specific parameters (ex: Add an element to $s$ that includes the sin of the angle between the pole and the cart (Cart pole simulation), we can further optimize the performance of LFA-NPG. As a guideline, we expect the optimal $\phi$ to have: 
    \begin{itemize}
        \item All elements of $s$ such that they cannot be derived from other elements within $s$ through simple operations
        \item All elements of $s$ exert sufficient influence over $V^{\pi _{\theta}}(\rho)$
    \end{itemize}.Therefore, an analysis of the environment and challenge is necessary to obtain the optimal $\phi$ for LFA-NPG. This is one of the fundamental distinctions between Linear function approximations of the value function, and Neural Network based approximations: Algorithms such as TRPO and PPO are capable of determining the optimal $\phi$ function themselves, ensuring that they maximize their potential. In complex, high dimensional challenges, this self selecting system is one of the greatest strengths of Neural Networks. However, we hypothesize that at lower dimensional systems, one could obtain the optimal $\phi$ through manual analysis. This places LFA-NPG on equal footing with TRPO and PPO with regards to $\phi$.

\section{Results and Discussion}

To determine the effectiveness of LFA-NPG in relation to TRPO and PPO, we test each algorithm on across 2 simulated environments (Figure 2): 
\begin{itemize}
    \item CartPole: Consider a system comprising of a cart that may move from side to side, with a joint connecting the cart to one end of a pole. The objective of our agent is to find the optimal policy so as to balance the pole atop the cart, by applying a force (left or right) to either the left or right\cite{barto83}. A reward of 1 is given for each timestep the pole is kept upwards, and after: a reward of 200.0 is reached; the pole falls over; or the cart goes out of bounds, the episode terminates.
    \item Acrobot: Consider a system featuring 2 links and 2 joints. Our objective is to swing the end of the lower link above a given height as fast as possible by applying torque (Clockwise, counterclockwise, or none) on the top link\cite{sutton96, geramifard15}. A reward of -1 is given for each episode where the bottom link is below our target height, and the episode terminates after either 500 timesteps, or after the bottom link reaches the goal height.
\end{itemize}

\begin{figure}%
    \centering
    \subfloat[\centering Cartpole-v0]{{\includegraphics[width=2cm]{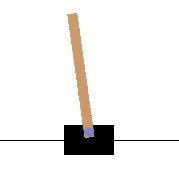} }}%
    \qquad
    \subfloat[\centering Acrobot-v1]{{\includegraphics[width=2cm]{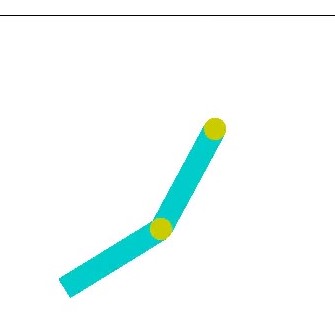} }}%
    \caption{Visualization of CartPole and Acrobot}%
    \label{fig:example}%
\end{figure}

The first step of testing begins with determining the optimal hyperparameters. The set of hyperparameters for LFA-NPG includes the NPG Weight $\eta$, SGD step size $\alpha$, Number of Actor iterations $T$, Number of Critic Iterations $R$, and the set of function approximators $\phi$. To determine the optimal set of function approximators to maximize $V^{\pi _{\theta}}(p)$, a through analysis of the problem is required. 
\begin{figure}[hbt!]
    \centering
    \subfloat[\centering Comparing $\phi$s for Cartpole]{{\includegraphics[width=5.5cm]{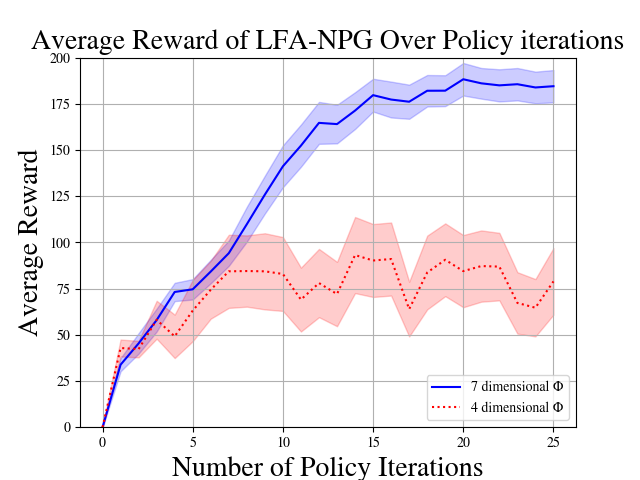} }}%
    \qquad
    \subfloat[\centering Comparing $\phi$s for Acrobot]{{\includegraphics[width=5.5cm]{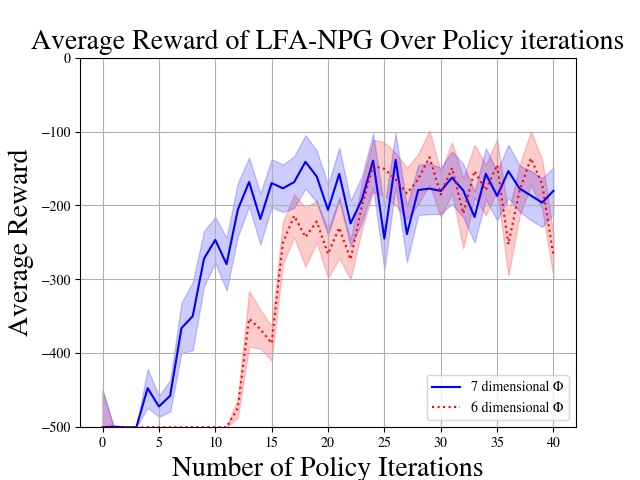} }}%
    \caption{Optimizing the set of function approximators $\phi$ for LFA-NPG}%
    \label{fig:example}%
\end{figure}
We begin with our analysis of the CartPole problem Our State $\mathcal{S} \in \mathbb{R}^4$ contains the position and velocity of the cart, along with the angle and angular velocity of the pole. We may find it fit to alter the dimensionality of $\mathcal{S}$ such that $\mathcal{S} \in \mathbb{R}^7$, where $\mathcal{S}$ now includes the position and velocity of the cart, sine and cosine of the angle of the pole, angular velocity of the pole, and sine and cosine of the angular velocity of the pole. We predict that this $\phi$ should offer a better result, as more features are learned each iteration, and we can observe a more complex relationship. We proceed to test LFA-NPG on CartPole, and compare the Reward per Policy Iteration to validate this assumption(Figure 3). For all future results of LFA-NPG in Cartpole, we use $\mathcal{S} \in \mathbb{R}^7$. We find our other hyperparameters as $T=25; N=150; \eta=0.1; \alpha=0.1; \mathcal{W}=10^{12}; \gamma=0.95$\footnotemark .
Analysis of Acrobot problem: Our initial State contains the sine and cosine of the two joint angles, and the angular velocities of each link. As $\mathcal{S} \in \mathbb{R}^6$, we have a sufficiently high dimensionality already. Therefore, we only add the sin of the angular velocity between the two joints to $\mathcal{S}$, such that $\mathcal{S} \in \mathbb{R}^7$. Our rationale is that in scenarios where the angle between the two joints is $\sim \pi$, we should observe a greater reward. We may characterize Acrobot as an sparse reward problem: The agent will get a constant minimum reward $(-500)$ until it explores and finds a greater reward. In acrobot, this occurs when our agent has a policy that returns reward $\mathcal{R}>-500$. Once more, we test LFA-NPG on Acrobot, and compare the performance for 6 and 7 dimensional $\phi$ (Figure 3). While after a greater number of policy iterations the reward for both $\phi$s are comparable, we note that $\mathcal{S} \in \mathbb{R}^7$ reaches these rewards after half the number of iterations. For all future results of LFA-NPG in Acrobot, we use $\mathcal{S} \in \mathbb{R}^7$. We find our other hyperparameters as $T=60; N=80; \eta=1; \alpha=0.0001; \mathcal{W}=10^{12}; \gamma=0.95$\footnotemark[\value{footnote}].
\footnotetext{Hyperparameters were experimentally determined}

We now proceed to compare the performance of LFA-NPG to that of flagship Neural Networks, TRPO and PPO\footnote{All tests were conducted on the same device; Intel i5-8250U CPU @1.60 GHz, 8 GB RAM, Intel UHD Graphics 620}.(Figure 4) 
\begin{figure}[hbt!]
    \centering
    \subfloat[\centering Cartpole: Reward vs Iterations]{{\includegraphics[width=5.5cm]{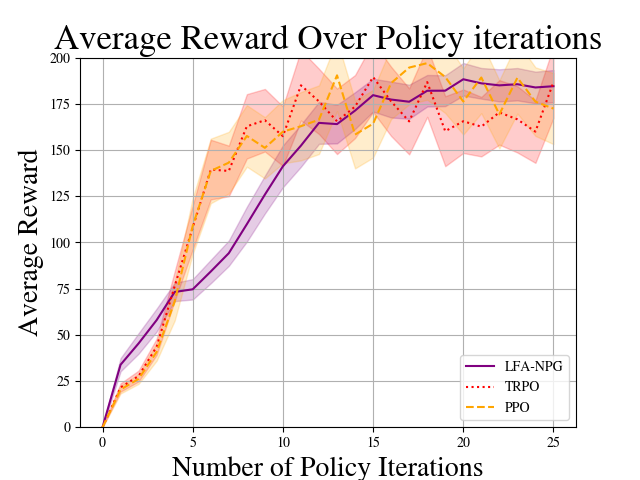} }}%
    \qquad
    \subfloat[\centering Cartpole: Processor Time (s) vs Reward]{{\includegraphics[width=5.5cm]{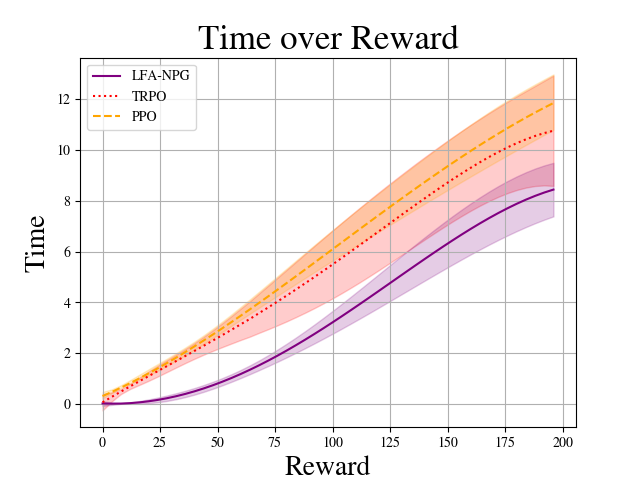} }}%
    \qquad
    \subfloat[\centering Acrobot: Reward vs Iterations]{{\includegraphics[width=5.5cm]{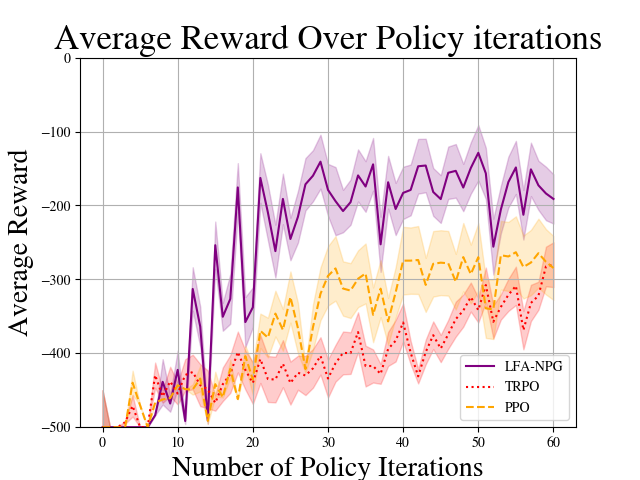} }}%
    \qquad
    \subfloat[\centering Acrobot: Processor Time (s) vs Iterations]{{\includegraphics[width=5.5cm]{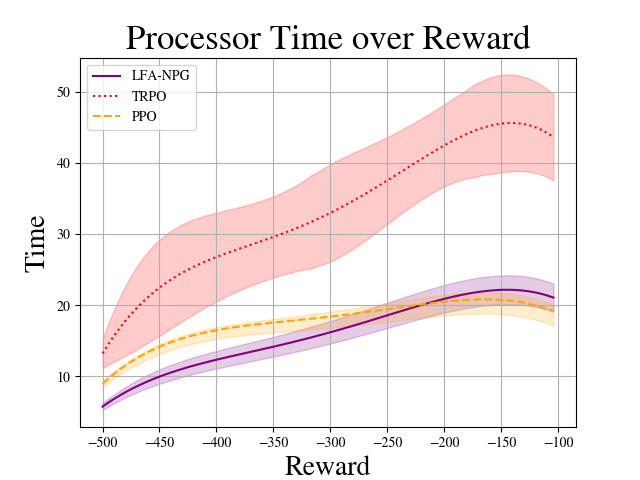} }}%
    \caption{Comparing LFA-NPG with Flagship Neural Networks}%
    \label{fig:example}%
\end{figure}

Having drawn comparisons between LFA-NPG, TRPO, and PPO, we now turn our attention to the task of analyzing the robustness of each algorithm. Robustness analysis involves sampling a State with some randomly sampled deviation of the true State, and observing how the performance of the algorithm is affected by this deviation. Mathematically we modify the returned State such that each element $S_k^n \rightarrow S_k^{n'}; S_k^{n'}= S_k^n \times z; z \sim (1+\zeta, 1-\zeta)$. We may then compare the effect of this Noise level $\zeta$ on the convergence and performance of algorithms (Figure 5).

\begin{figure}[p!]
    \centering
    \subfloat[\centering Cartpole: LFA-NPG Reward vs Policy Iterations]{{\includegraphics[width=5.5cm]{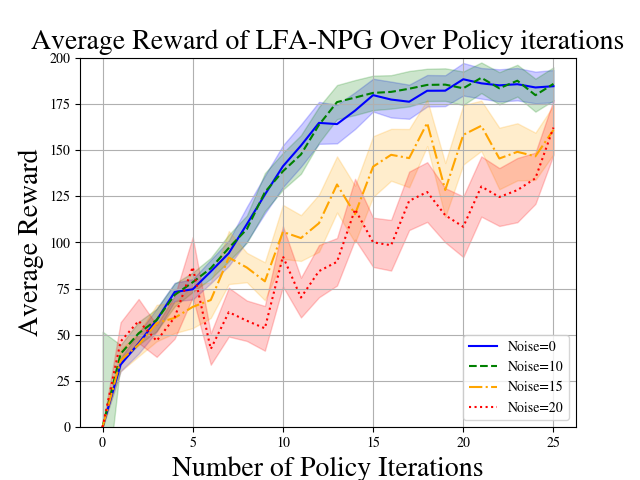} }}%
    \qquad
    \subfloat[\centering Cartpole: TRPO Reward vs Policy Iterations]{{\includegraphics[width=5.5cm]{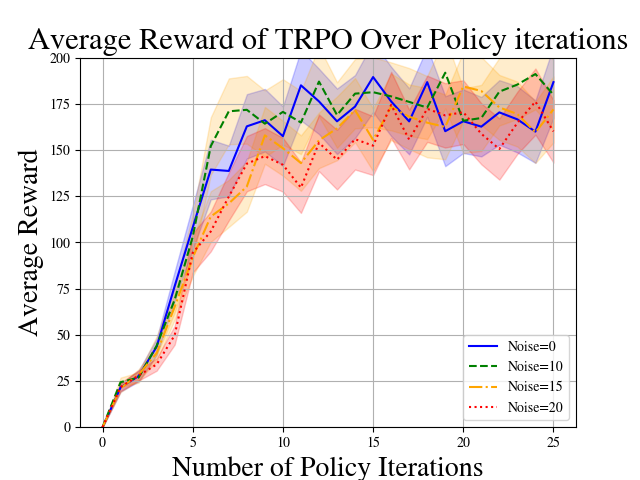} }}%
    \qquad
    \subfloat[\centering Cartpole: PPO Reward vs Policy Iterations]{{\includegraphics[width=5.5cm]{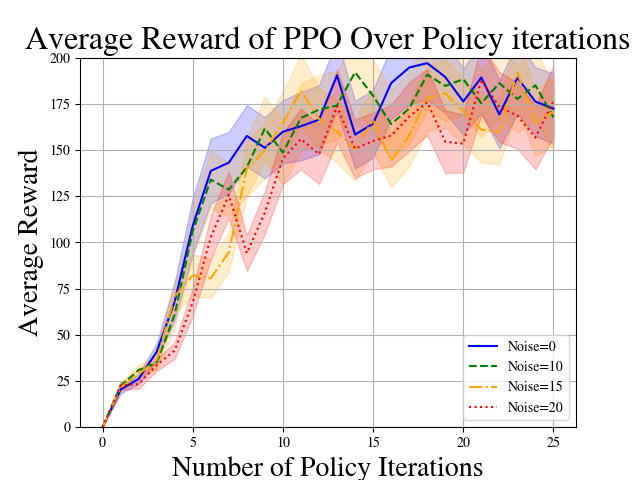} }}%
    \qquad
    \subfloat[\centering Acrobot: LFA-NPG Reward vs Policy Iterations]{{\includegraphics[width=5.5cm]{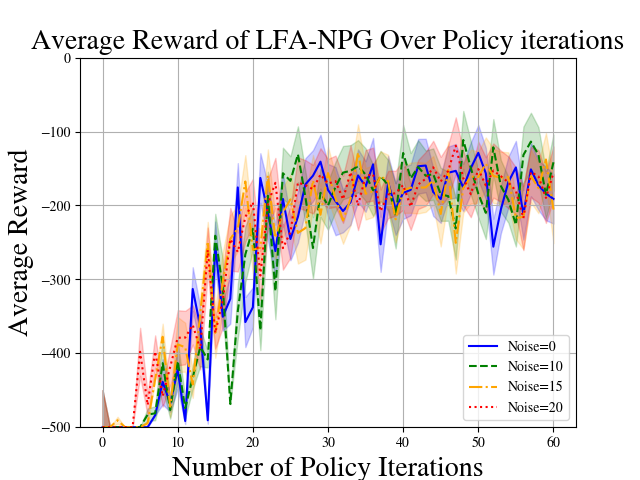} }}%
    \qquad
    \subfloat[\centering Acrobot: TRPO Reward vs Policy Iterations]{{\includegraphics[width=5.5cm]{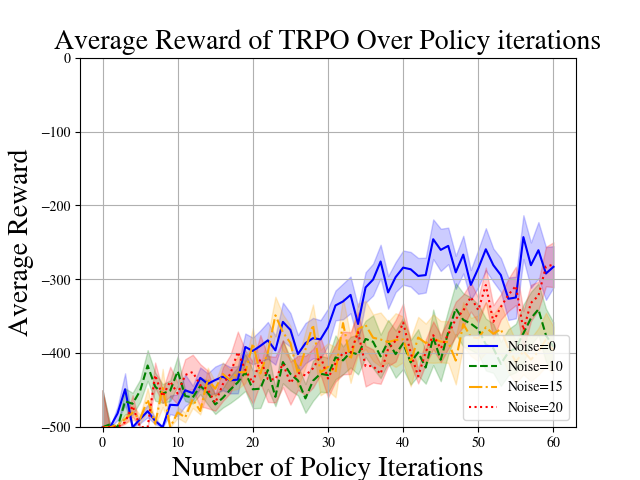} }}%
    \qquad
    \subfloat[\centering Acrobot: PPO Reward vs Policy Iterations]{{\includegraphics[width=5.5cm]{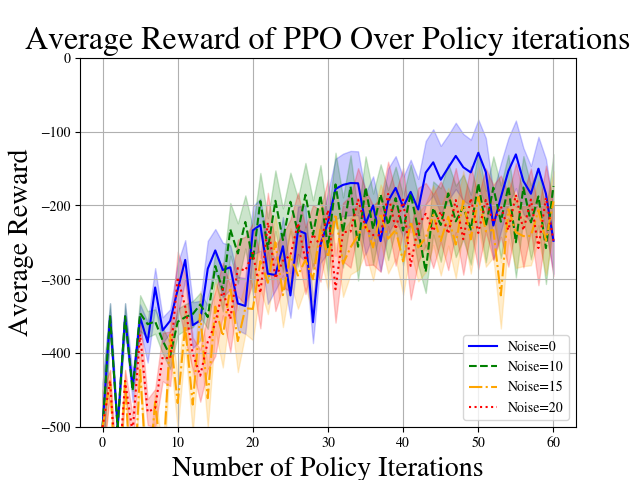} }}%
    \caption{LFA-NPG, TRPO, and PPO Robustness Analysis}%
    \label{fig:example}%
\end{figure}

\section{Conclusion}

    In this work, we compare our LFA-NPG Natural Actor Critic algorithm with the 2 main flagship RL Algorithms, TRPO and PPO. Intuitively, we expected that with lower dimensional discrete challenges such as CartPole and Acrobot, LFA-NPG should attain a similar Result as TRPO and PPO, and we also hypothesize that LFA-NPG should reach each reward in a shorter duration of time. In the cartpole simulation, we find this hypothesis to be wholly supported: We are successfully able to demonstrate that LFA-NPG outperforms the leading RL algorithms, and is able to achieve each reward in  time 20\%+ faster than either TRPO or PPO. This makes intuitive sense, as the natural gradient architecture of LFA-NPG is significantly less computationally expensive as a complex Neural Network architecture, and therefore can be executed much swifter. 
    
    When testing these same algorithms on the sparse reward environment of Acrobot, we observe LFA-NPG noticeably outperforms both PPO and TRPO, and we also observe LFA-NPG and PPO acheiving a similar speed, which is significantly faster than that of TRPO. We believe that LFA-NPG is able to drastically outperform these more complex architectures due to the nature of an sparse reward challenge: Specifically, that the nature of Natural Gradient Descent means that it will take large steps if it is not able to find the optimal result. This enables it to find a policy that works faster than TRPO or PPO, which are constrained by the KL-Divergence. We also observe a distinction in performance between TRPO and PPO, where PPO is able to attain a better result. We attribute this to the nature of PPOs multi-batch updates, which allow it to ensure that it improves its reward after each successive step, enabling it to improve its performance with swift updates. The difference in performance between PPO and TRPO may also be attributed to overfitting, where TRPO may be too cautious in each update to converge swiftly.
    
    We also draw conclusions on the robustness of each algorithm. In CartPole, We observe that LFA-NPG is wholly resistant to $\zeta<10$, and exhibits more variance and gains less reward with more variance. TRPO and PPO both have similar levels of noise resistant, and gain a similar reward with larger levels of noise. In the acrobot simulation, we observe that LFA-NPG is very noise resistant, exhibiting effectively the same performance across a wide spread of $\zeta$. TRPO has an initial drop in performance once $\zeta \neq 0$, and PPO has a very gradual decline in performance as $\zeta$ increases. With this in mind, we may summarize the results of our research as follows:
\begin{itemize}
    \item In standard low dimensional challenges, Natural Policy Gradient Methods such as LFA-NPG reach optimal rewards in noticeably less time than State of the art Complex Neural Networks, while maintaining a similar rate of convergence
    \item In sparse reward low dimensional challenges, Natural Policy Gradient Methods converge significantly faster than State of the art Complex Neural Networks, while achieving the same or greater speed
\end{itemize}
This allows us to conclude that in a challenge with sufficiently small State and action space, Natural Policy Gradients are a better choice than the leading neural networks in RL, validating our initial assumption. When combining this with their comparative simplicity to practically implement, we demonstrate that algorithms such as LFA-NPG are very effective for a multitude of RL Use Cases, and often surpass their more complex brethren. The potential applications of LFA-NPG range from classic control applications from RL Literature to backend applications within more complex systems (such as optimizing a function to control the velocity of the drone). 
In the future, we would hope to formulate a version of LFA-NPG that can find the optimal policy for a continuous action space. This necessitates the formulation of a $\phi$ function that includes the action within the matrix. We also hope to formulate a form of LFA-NPG that has a method to self select the optimal $\phi$, through some linear method.

\newpage
\bibliography{mybib}{}

\begin{thebibliography}{10}

\bibitem{kiran2021deep}
B.~R. Kiran, I.~Sobh, V.~Talpaert, P.~Mannion, A.~A.~A. Sallab, S.~Yogamani,
  and P.~Pérez, ``Deep reinforcement learning for autonomous driving: A
  survey,'' 2021.

\bibitem{openai2019solving}
OpenAI, I.~Akkaya, M.~Andrychowicz, M.~Chociej, M.~Litwin, B.~McGrew,
  A.~Petron, A.~Paino, M.~Plappert, G.~Powell, R.~Ribas, J.~Schneider,
  N.~Tezak, J.~Tworek, P.~Welinder, L.~Weng, Q.~Yuan, W.~Zaremba, and L.~Zhang,
  ``Solving rubik's cube with a robot hand,'' 2019.

\bibitem{BHATNAGAR20092471}
S.~Bhatnagar, R.~S. Sutton, M.~Ghavamzadeh, and M.~Lee, ``Natural
  actor–critic algorithms,'' {\em Automatica}, vol.~45, no.~11,
  pp.~2471--2482, 2009.

\bibitem{PETERS20081180}
J.~Peters and S.~Schaal, ``Natural actor-critic,'' {\em Neurocomputing},
  vol.~71, no.~7, pp.~1180--1190, 2008.
\newblock Progress in Modeling, Theory, and Application of Computational
  Intelligenc.

\bibitem{khodadadian2021finite}
S.~Khodadadian, T.~T. Doan, S.~T. Maguluri, and J.~Romberg, ``Finite sample
  analysis of two-time-scale natural actor-critic algorithm,'' 2021.

\bibitem{haarnoja2019soft}
T.~Haarnoja, A.~Zhou, K.~Hartikainen, G.~Tucker, S.~Ha, J.~Tan, V.~Kumar,
  H.~Zhu, A.~Gupta, P.~Abbeel, and S.~Levine, ``Soft actor-critic algorithms
  and applications,'' 2019.

\bibitem{xu2021graphembedded}
X.~Xu, Q.~Chen, X.~Mu, Y.~Liu, and H.~Jiang, ``Graph-embedded multi-agent
  learning for smart reconfigurable thz mimo-noma networks,'' 2021.

\bibitem{schulman2017trust}
J.~Schulman, S.~Levine, P.~Moritz, M.~I. Jordan, and P.~Abbeel, ``Trust region
  policy optimization,'' 2017.

\bibitem{schulman2017proximal}
J.~Schulman, F.~Wolski, P.~Dhariwal, A.~Radford, and O.~Klimov, ``Proximal
  policy optimization algorithms,'' 2017.

\bibitem{roberts21}
D.~A. Roberts, S.~Yaida, and B.~Hanin, ``The principles of deep learning
  theory,'' 2021.

\bibitem{bartosutton}
R.~S. Sutton and A.~G. Barto, {\em Reinforcement learning: An introduction}.
\newblock MIT press, 2018.

\bibitem{agarwal2020theory}
A.~Agarwal, S.~M. Kakade, J.~D. Lee, and G.~Mahajan, ``On the theory of policy
  gradient methods: Optimality, approximation, and distribution shift,'' 2020.

\bibitem{barto83}
R.~S. AG~Barto and C.~Anderson, ``Neuronlike adaptive elements that can solve
  difficult learning control problem,'' 1983.

\bibitem{sutton96}
R.~Sutton, ``Generalization in reinforcement learning: Successful examples
  using sparse coarse coding,'' 2015.

\bibitem{geramifard15}
A.~Geramifard, C.~Dann, R.~H. Klein, W.~Dabney, and J.~P. How, ``Rlpy: A
  value-function-based reinforcement learning framework for education and
  research,'' {\em Journal of Machine Learning Research}, vol.~16, no.~46,
  pp.~1573--1578, 2015.

\end{thebibliography}
\bibliographystyle{ieeetr} 

\end{document}